%% file: main.tex
\documentclass[Afour,sageh,times]{sagej}

\usepackage{moreverb,url}
\usepackage{multirow}
\usepackage{array}
\usepackage[colorlinks,bookmarksopen,bookmarksnumbered,citecolor=blue,urlcolor=blue,linkcolor=blue]{hyperref}

\definecolor{placeholdergray}{gray}{0.45}

\def\volumeyear{2026}

\makeatletter
\def\@maketitle{%
\vspace*{-34pt}\null
\begin{center}\begin{sf}
\begin{minipage}[t]{\textwidth}
  \vskip 12.5pt
  {\raggedright\titlesize\textbf{\@title}\par}
  \vskip 0.5em
\end{minipage}
\vskip 1.2em
{\par\large\lineskip .5em{\raggedright\textbf{\@author}\par}}
\vskip 40pt
{\noindent\usebox\absbox\par}
{\vspace{20pt}{\noindent\normalsize\@keywords}\par}
\end{sf}\end{center}
\vspace{22pt}\par}
\def\ps@title{%
  \def\@oddhead{\parbox{\textwidth}{\mbox{}\\[-1pt]\noindent\rule{\textwidth}{0.5pt}}}%
  \let\@evenhead\@oddhead
  \def\@oddfoot{\parbox[t]{\textwidth}{\scriptsize\itshape Preprint, \today}}%
  \let\@evenfoot\@oddfoot}
\def\ps@sagepage{%
  \let\@mkboth\@gobbletwo
  \def\@evenhead{\parbox{\textwidth}{\normalsize\sagesf\thepage\hfill\itshape Rosetta Touch (preprint)\\[-6pt]%
    \noindent\rule{\textwidth}{0.25pt}}}%
  \def\@oddhead{\parbox{\textwidth}{\normalsize\sagesf{\itshape{\leftmark}}\hfill\thepage\\[-6pt]%
    \noindent\rule{\textwidth}{0.25pt}}}%
  \def\@evenfoot{}\def\@oddfoot{}}
\makeatother
\begin{document}

\runninghead{Hou et~al.}

\title{Rosetta Touch: a force-matched tactile dataset across sensors}

\author{Jingchao Hou\affilnum{1,2}, Gabriele M. Caddeo\affilnum{1}, Pasquale Marra\affilnum{1},
Francesco Donato\affilnum{1}, Raffaello Camoriano\affilnum{2} and Lorenzo Natale\affilnum{1}}

\affiliation{\affilnum{1}Humanoid Sensing and Perception, Istituto Italiano di Tecnologia, Genova, Italy\\
\affilnum{2}Politecnico di Torino, Torino, Italy}

\corrauth{Gabriele M. Caddeo, Istituto Italiano di Tecnologia, Center for Robotics and Intelligent Systems,
Via San Quirico 19D, 16163 Genova, Italy.}

\email{gabriele.caddeo@iit.it}

\input{sections/00_abstract}


\maketitle

\input{sections/01_introduction}
\input{sections/02_related_work}
\input{sections/03_collection}
\input{sections/04_dataset}

\input{sections/05_validation}
\input{sections/06_next_steps}

\begin{acks}{We thank Daimon Robotics for their support.}
\end{acks}

\bibliographystyle{SageH}
\bibliography{references}

\end{document}

%% file: sections/00_abstract.tex
\begin{abstract}
Tactile representations and policies learned on one sensor rarely survive a change of sensor, or even a change of
unit of the same sensor model, because each device turns the same contact into a different signal. Learning a
representation that bridges sensors needs data in which different sensors record the same contact under the same
force, and existing tactile datasets do not provide it. We announce Rosetta Touch, a dataset that does: like the
Rosetta Stone, which carries one text in several scripts, it records the same contacts on several tactile sensors,
with the applied force matched across them. An admittance-controlled delta robot with a six-axis force/torque sensor presses indenters into each sensor
along identical, seeded force trajectories, so that force, contact location, indenter geometry and contact phase are
shared across sensors while each gel's own response is not. The current release covers three optical sensor families
of different design (DIGIT, GelSight Mini and Daimon), 23 indenters and 945 force trajectories per indenter on one
unit of each, about 65,000 force-matched presses in total, plus held-out recordings on four further units. A
preliminary encoder trained on these pairs reads force across sensors as well as within a sensor on objects it never
saw, but does not transfer to new units of DIGIT. Collection continues with a fourth, magnetic family (XELA), two units
per family, about 50 indenters, and force trajectories modelled on manipulation.
\end{abstract}

%% file: sections/01_introduction.tex
\section{Introduction}\label{sec:intro}

\begin{figure*}[t]
\centering
\includegraphics[width=\textwidth]{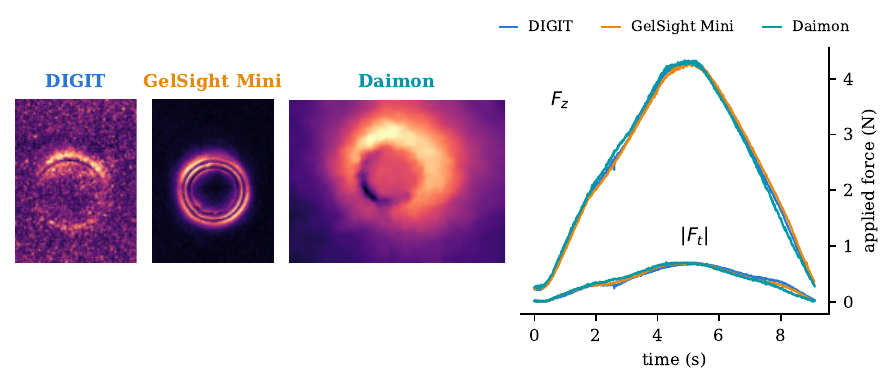}
\caption{One press recorded on three sensors: the same trial of the ring indenter at the centre of the contact area,
during the hold at 4.3\,N normal and 0.7\,N tangential force. Left: the change from the trial's no-contact image
(DIGIT, GelSight Mini) and the magnitude of the in-plane deformation (Daimon), full frames. Right: the force measured by
the robot's force/torque sensor in the three recordings of this trial, which follow the same commanded history.}
\label{fig:teaser}
\end{figure*}

Contact-rich manipulation needs information that vision cannot provide: contact forces, local geometry and the
state of the interface \citep{romano2011human,lee2019making,tian2019manipulation}. Vision-based tactile sensors
\citep{yuan2017gelsight,lambeta2020digit,wardcherrier2018tactip} have become the dominant way to obtain it, and
large tactile models \citep{higuera2024sparsh,feng2025anytouch,feng2026anytouch2} now learn from their images at
scale. Yet a representation or a policy trained on one sensor rarely transfers to another. Illumination, elastomer
compliance and coating, and pad geometry differ between designs, and manual fabrication makes even two units of
the same model differ \citep{shahidzadeh2025feelanyforce, 10851808}. Moving a tactile policy to a new sensor today means
recollecting demonstrations.

Learning across sensors needs data in which different sensors record the same contact, and existing datasets do not
provide it. Large collections pair touch with vision or language but not with force. Force-labelled collections were
recorded one sensor at a time \citep{higuera2024sparsh,shahidzadeh2025feelanyforce}; multi-sensor collections align
sensors by contact location, either without force \citep{feng2025anytouch} or with force measured but not imposed, so
the force history at a location differs between sensors \citep{chen2025general,feng2026anytouch2}. In all of them each
sensor model is represented by a single unit (Section~\ref{sec:related}).

Our key observation is that \emph{force} is the quantity through which manipulation is controlled, and the one that
can be made identical across sensors. Humans adapt the balance between grip and load forces to the friction of the surface they
hold, and lose this ability when the fingertips are anaesthetised \citep{johansson1984roles,johansson2009coding}; robot manipulation
in contact is formulated as the control of interaction forces \citep{hogan1985impedance,suomalainen2022survey}; and
manipulation policies improve markedly when they sense or command force
\citep{he2025foar,liu2025forcemimic,marra2026multifingered,marra2026temporal}. An admittance-controlled robot can reproduce the same force history on every sensor to
within a few hundredths of a newton, so force, contact location, indenter geometry and contact phase become shared
across sensors, while the indentation depth remains a property of each gel. Every frame of one sensor can then be paired with the frames of every
other sensor that recorded the same physical state (Figure~\ref{fig:teaser}).

This preprint announces Rosetta Touch at its current stage and the collection that is under way. The name refers to
the Rosetta Stone, which carries the same decree in three scripts and so made it possible to read one script through
another; Rosetta Touch carries the same contacts in the signals of several tactile sensors, so that one sensor can be
read through another.
\begin{itemize}
\item \textbf{Recorded:} about 65,000 force-matched presses of 23 indenters on one unit each of DIGIT, GelSight Mini
and Daimon, three optical sensors of different design, plus held-out recordings on four further units
(Sections~\ref{sec:collection}--\ref{sec:dataset}).
\item \textbf{A first use:} an encoder trained on the pairs reads force across sensors as well as within a sensor on
unseen objects, and fails on new DIGIT units, which shapes the rest of the collection (Section~\ref{sec:validation}).
\item \textbf{Under way:} a fourth, magnetic family (XELA), two units per family, about 50 indenters and force
trajectories modelled on manipulation (Section~\ref{sec:next}).
\end{itemize}

%% file: sections/02_related_work.tex
\section{Related work}\label{sec:related}

\begin{table*}[t]
\footnotesize\sf\centering
\setlength{\tabcolsep}{4pt}
\caption{Force-labelled and multi-sensor tactile datasets. Units: physical units of the same sensor model;
n/r: not reported.}
\label{tab:datasets}
\begin{tabular}{>{\raggedright\arraybackslash}p{0.11\textwidth}>{\raggedright\arraybackslash}p{0.17\textwidth}>{\raggedright\arraybackslash}p{0.115\textwidth}>{\raggedright\arraybackslash}p{0.15\textwidth}>{\raggedright\arraybackslash}p{0.13\textwidth}>{\raggedright\arraybackslash}p{0.095\textwidth}>{\raggedright\arraybackslash}p{0.085\textwidth}}
\toprule
Dataset & Sensors (units) & Force ground truth & Contacts & Alignment across sensors & Size & Public\\
\midrule
Sparsh force & DIGIT, GelSight (units n/r) & ATI Nano17 & 3 indenters; normal load $\le$5\,N, then 2\,mm slide & none (separate recordings) & 75k samples & DIGIT part\\
FeelAnyForce & GelSight Mini (1 for training; 3 for evaluation); DIGIT tested & 3-axis F/T (0.04\,N resolution) & 10 indenters; presses at sampled poses & none & $>$200k indentations & yes\\
TacQuad & GelSight Mini, DIGIT, DuraGel, Tac3D (units n/r) & none & 124 objects; presses & same position and speed & 72.6k frames & yes\\
GenForce & 5 in-house GelSight variants; GelSight-type, TacTip, uSkin & ATI Nano17 & normal and shear load / unload & same contact location & 100--180k images per sensor & not found\\
UniForce & in-house GelSight, TacTip, uSkin (1 each) & none (equilibrium) & 8 indenters; hand-guided drag / rotate in a gripper & simultaneous two-finger contact & $\sim$8k frames per pair & announced\\
ToucHD-Force & GelSight Mini, DIGIT, DuraGel (+2 recorded, unused) & 6-axis F/T at the wrist (measured) & 71 indenters; press and slide in 4 directions & coarse (nearby locations) & 722k pairs & yes\\
\midrule
Ours (current) & DIGIT, GelSight Mini, Daimon (1 unit each; +4 test units); XELA and 2 units per family under way & ATI Nano17; force commanded (admittance) & 23 indenters (50 planned); load--hold--unload & same commanded force history; paired at matched force and phase & 65.8k trials (partial) & planned\\
\bottomrule
\end{tabular}
\end{table*}

\subsection{Tactile datasets without force}
Most large tactile datasets pair touch with other modalities rather than with force. VisGel \citep{li2019connecting}
and Touch and Go \citep{yang2022touch} pair GelSight images with vision, ObjectFolder \citep{gao2021objectfolder} adds
audio and object models, TVL \citep{fu2024touch} and Octopi \citep{yu2024octopi} add language, and YCB-Slide
\citep{suresh2023midastouch} records DIGIT sliding over objects with poses. They drive representation learning
\citep{higuera2024sparsh,feng2025anytouch} but carry no force ground truth and, being recorded with one sensor each,
no correspondence between sensors.

\subsection{Force-labelled and multi-sensor datasets}
Table~\ref{tab:datasets} lists the datasets closest to ours. Force-labelled collections were first recorded on one
sensor at a time: the force data of Sparsh \citep{higuera2024sparsh} (DIGIT and GelSight, three indenters, a normal
load followed by a 2\,mm slide) and FeelAnyForce \citep{shahidzadeh2025feelanyforce} (over 200k indentations on a
GelSight Mini with ten indenters, transferred to further units by calibration). TacQuad \citep{feng2025anytouch}
records four sensors pressing the same position of the same object at the same speed, aligning contacts by location,
without force. GenForce \citep{chen2025general} labels force on several in-house GelSight variants and on a
heterogeneous set (GelSight-type, TacTip, uSkin) and transfers force labels between sensors through contacts at the
same location. UniForce \citep{chen2026uniforce} avoids force sensors altogether: two sensors on a parallel gripper
squeeze an object and are assumed to carry equal and opposite forces. ToucHD-Force \citep{feng2026anytouch2} is the
largest: 722k touch--force pairs recorded while 71 indenters press and slide on several optical sensors, with the
force measured by a wrist force/torque sensor.

In all of them the force is measured, not imposed, so two sensors are at best aligned by contact location; the force
history at that location differs from sensor to sensor because the gels differ. Each sensor model is also represented
by a single physical unit, or by in-house variants, so the variability between units of the same commercial model is
not captured. In our dataset the same force history is commanded on every sensor, so frames can be paired at matched
force and contact phase; it already includes test recordings on further units of each model, and the ongoing
collection records two units of every family and adds a magnetic taxel sensor alongside the optical ones.

\subsection{Cross-sensor tactile representations}
Methods that learn representations shared across sensors are limited by this data. T3 \citep{zhao2024transferable},
Sparsh \citep{higuera2024sparsh}, AnyTouch and AnyTouch~2 \citep{feng2025anytouch,feng2026anytouch2} learn from
unpaired or coarsely aligned collections; SITR \citep{gupta2025sensor} aligns geometry through simulation; TACTX
\citep{park2026tactx} and UniTac-VAE \citep{hou2026unitac} map sensors into shared latent spaces; TransForce
\citep{chen2025transforce}, GenForce \citep{chen2025general} and UniForce \citep{chen2026uniforce} target force transfer
specifically. None of them can be evaluated on whether a read-out transfers between sensors at the same force, or between two units
of the same model; force-matched pairs on several units make both measurable.

%% file: sections/03_collection.tex
\section{Collection}\label{sec:collection}

\begin{figure}[t]
\centering
\includegraphics[width=\linewidth]{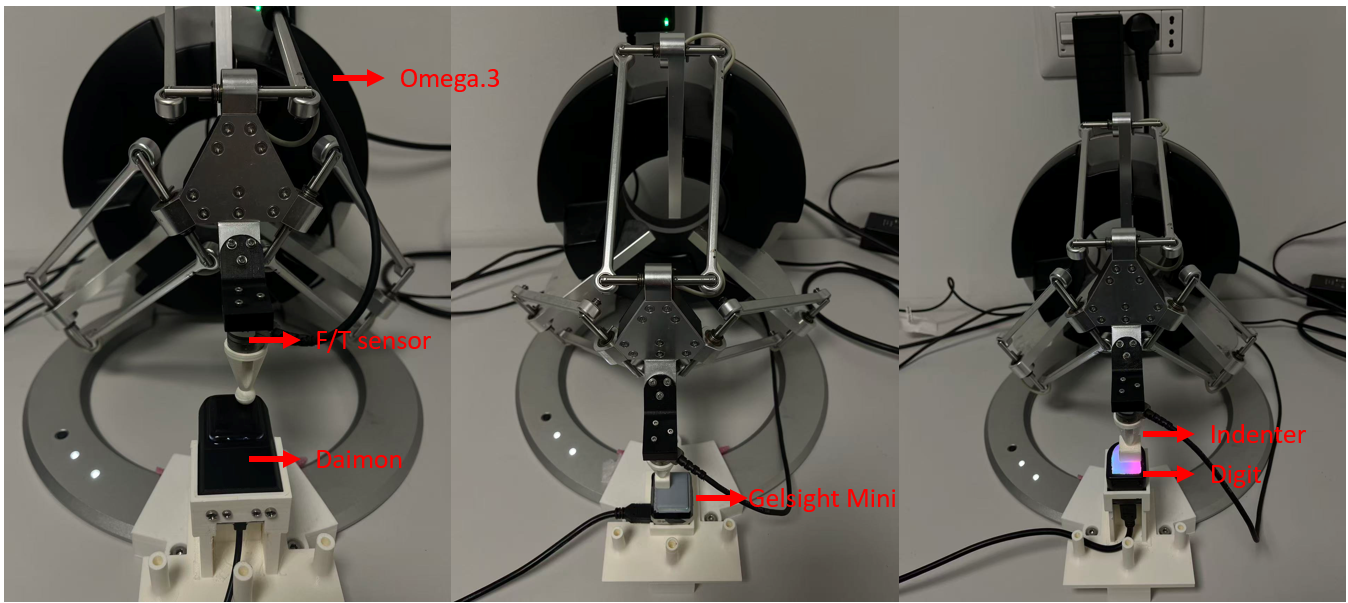}
\caption{The collection rig with Daimon, GelSight Mini and DIGIT (left to right).}
\label{fig:rig}
\end{figure}

\paragraph{Platform.}
An omega.3 haptic device (Force Dimension), a delta mechanism with three translational degrees of freedom, moves the
indenter (Figure~\ref{fig:rig}). An ATI Nano17 force/torque sensor between the end effector and the indenter measures
the force applied to the tactile sensor at about 770\,Hz. Each tactile sensor sits on a 3D-printed support at the
centre of the robot's workspace; three reference pillars on the support and a known pillar-to-tip offset on every
indenter give the transformation from the robot frame to the sensor frame, so contact points and forces are expressed
in the sensor frame. All streams are timestamped on a shared network clock.

\paragraph{Sensors.}
DIGIT \citep{lambeta2020digit} and GelSight Mini \citep{yuan2017gelsight} image an illuminated elastomer with a camera
(RGB, 320$\times$240, recorded at 30 and 20\,Hz). Daimon (DM-Tac WM, \citealp{daimon}) is also optical but its SDK
outputs a deformation field: in-plane displacement and depth in micrometres on a 240$\times$320 grid, recorded at
30\,Hz. The three differ in illumination, elastomer and pad size (17.4$\times$13.2, 18.8$\times$14.6 and
32.0$\times$24.0\,mm). Contacts are placed in the usable area of the smallest pad (DIGIT), centred on every sensor.

\begin{figure*}[t]
\centering
\includegraphics[width=\textwidth]{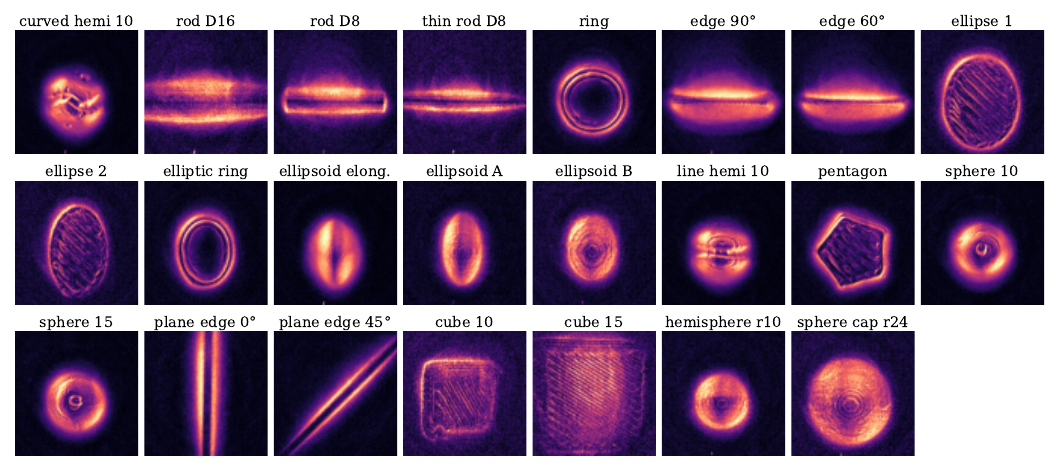}
\caption{The 23 indenters, each as imprinted on GelSight Mini: change from the no-contact image during the hold of a
press at the centre of the contact area (normal force 2.3--8\,N).}
\label{fig:indenters}
\end{figure*}

\paragraph{Indenters.}
23 3D-printed indenters cover spheres and spherical caps of several radii, ellipsoids, rods, flat faces, edges at
several angles, rings and a pentagon. Figure~\ref{fig:indenters} shows each of them as GelSight Mini sees it; their CAD
files are part of the release.

\begin{figure}[t]
\centering
\includegraphics[width=\linewidth]{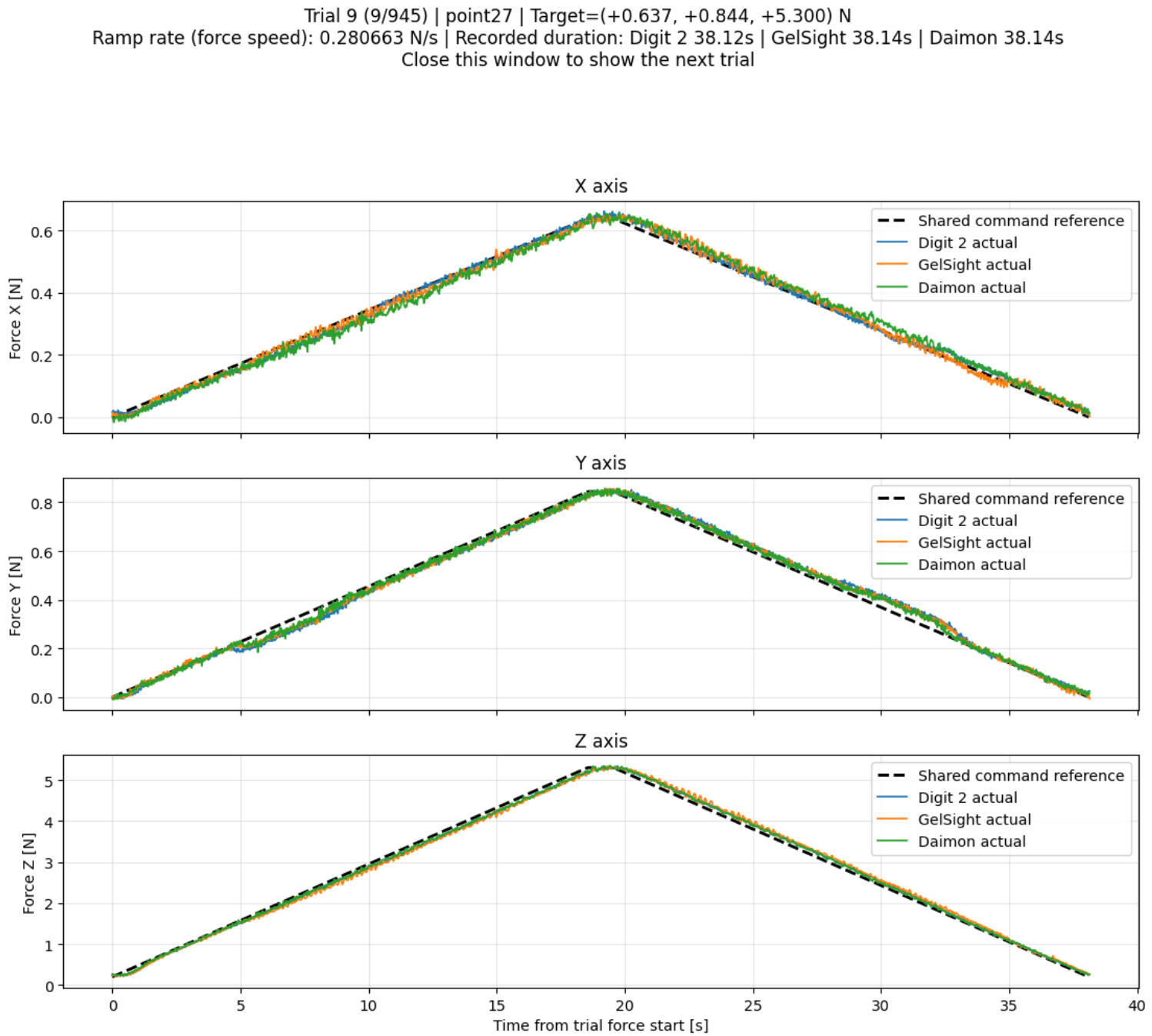}
\caption{Force tracking on the three sensors for one shared commanded trajectory (dashed): $x$, $y$ and $z$
components.}
\label{fig:tracking}
\end{figure}

\paragraph{Force control.}
The robot runs an admittance controller,
$M\ddot{x} + K_d\dot{x} = F^{*}(t) - F_{\mathrm{meas}}(t)$, with $M = \mathrm{diag}(15,15,10)$\,kg and
$K_d = \mathrm{diag}(350,350,550)$\,N\,s\,m$^{-1}$, so the indenter follows a commanded force history $F^{*}(t)$
whatever the compliance of the gel underneath. Position control could not do this: the same indentation depth gives
different forces on different gels. At steady state the force error stays within $\pm0.025$\,N tangentially and
$\pm0.08$\,N normally, and the three sensors follow the same commanded history closely during ramps
(Figure~\ref{fig:tracking}).

\paragraph{Common force envelope.}
For every indenter and sensor, a force-cone test raises the tangential force at fixed normal force until the indenter
slips. The largest tangential force that no sensor slips under, $F_{t,\max}(F_z) = \min_s F^{(s)}_{t,\max}(F_z)$,
bounds all commanded forces, with a safety factor of 0.8. No trial therefore slips on any sensor, and every trial can
be executed identically on all of them.

\paragraph{Force plans.}
Each trial loads linearly from a 0.2\,N home force to a target force, holds it for 1\,s and unloads. Targets are
sampled uniformly over the envelope disk at normal forces from 0.3 up to 8\,N (indenter-dependent), in random
tangential directions, with ramp rates drawn between 0.2 and 1.5\,N\,s$^{-1}$; targets that push outward near the edge
of the contact area are resampled. Each indenter gets 63 contact points on a 2\,mm grid and 15 force trajectories per
point, 945 trials. Plans are seeded by indenter and trial index, so every sensor executes the same 945 trials.

%% file: sections/04_dataset.tex
\section{The dataset}\label{sec:dataset}

\paragraph{Contents.}
Table~\ref{tab:contents} lists what is recorded. On the main unit of each sensor every indenter has all 945 trials. Four further units (two more DIGITs, a second GelSight Mini and a second Daimon) were recorded on three indenters with their own
seeded plan (5 contact points $\times$ 10 force trajectories), and are kept as a test set for transfer to new units. We recorded more DIGIT units because DIGIT units are known to be less consistent with each other \citep{10851808}.
The raw data amount to about 3\,TB.

\begin{table}[t]
\small\sf\centering
\caption{Recorded so far.}
\label{tab:contents}
\begin{tabular}{llrr}
\toprule
Sensor & Unit & Indenters & Trials\\
\midrule
DIGIT & 1 & 23 & 21,735\\
GelSight Mini & 1 & 23 & 21,735\\
Daimon & 1 & 23 & 21,735\\
\midrule
DIGIT & 2, 3 & 3 & 150 each\\
GelSight Mini & 2 & 3 & 150\\
Daimon & 2 & 3 & 150\\
\midrule
Total & 7 units & & 65,805\\
\bottomrule
\end{tabular}
\end{table}

\paragraph{Per run.}
Each run (one sensor unit and one indenter) contains the seeded force plan, the run configuration, the force/torque
stream with the applied force in the sensor frame and the trial phase, the robot state, and the tactile frames with
their timestamps: PNG images for DIGIT and GelSight Mini, and NumPy arrays of deformation and depth in micrometres for
Daimon. Every trial also records a no-contact frame during the approach, which serves as the reference for that trial.

\paragraph{Labels and pairing.}
Every frame inherits the applied force interpolated from the force/torque stream, the contact point in the sensor
frame and in a frame common to all sensors, the indenter, and the trial phase (loading, hold, unloading). Because all
sensors run the same plan, frames of two sensors can be paired within a trial: on a time grid of the slowest sensor,
each frame is matched to the frame of the other sensor with the nearest measured force in the same phase. Pairs never
cross from loading to unloading, since gel hysteresis gives a different image at the same force on the way down.
Over the 1.6 million pairs built so far (13 indenters), the applied force of paired frames differs by a median of
0.012--0.013\,N in normal force and 0.022--0.027\,N in tangential force (95th percentile of the total difference:
0.10--0.11\,N).

%% file: sections/05_validation.tex
\section{A first use: one embedding for three sensors}\label{sec:validation}

To check that the pairs are usable, we trained an encoder on an earlier subset of the data (9 indenters on the three
main units). A pretrained tactile trunk \citep{feng2026anytouch2}, adapted with low-rank updates, feeds a head that maps
8-frame windows of any of the three sensors into one 256-dimensional embedding. Training pulls together frames of
two sensors recorded at matched force and phase, and reads force, contact location and phase from the embedding with
heads shared by all sensors. We then fit linear read-outs on one sensor and test them on the same sensor
(\emph{within}) or on another (\emph{cross}).

\begin{table}[t]
\small\sf\centering
\caption{Linear read-outs ($R^2$) from the embedding, within a sensor. Gap: largest difference between a read-out
fitted on another sensor and the sensor's own.}
\label{tab:validation}
\begin{tabular}{lllr}
\toprule
Indenters & Quantity & Within & Gap\\
\midrule
Seen (new trials) & force & 0.94--0.99 & 0.007\\
Unseen & $F_z$ & 0.92--0.99 & 0.011\\
Unseen & shear, cameras & 0.76--0.84 & 0.011\\
Unseen & shear, Daimon & 0.97--0.99 & 0.011\\
\bottomrule
\end{tabular}
\end{table}

On indenters never used in training, a read-out fitted on one sensor reads another sensor as well as that sensor's
own read-out (Table~\ref{tab:validation}); a classifier identifies the sensor from the embedding 41\% of the time
(chance 33\%), against 100\% from the pretrained trunk's features. The paired data therefore produce an embedding in
which force is read the same way on all three sensors.

The held-out units show where this stops. A second Daimon unit is read as well as the training unit. A second
GelSight Mini keeps normal force, location and phase but loses part of the shear direction ($F_x$: 0.72 against 0.87).
Two further DIGIT units fail (normal force $R^2$ 0.62 and $-0.02$ against 0.95), although their frames carry the
information: before paired training, a read-out fitted on a new DIGIT's own frames reads its shear as well as one
fitted on the training unit's frames. For DIGIT, an encoder trained on one unit learns that unit rather than the sensor model, a known generalisation
problem of DIGIT \citep{10851808}: the encoder needs more than one DIGIT unit to learn the model, which is why every
family is now recorded on two units. GelSight Mini and Daimon units are much more consistent with each other, so for
them the second unit is a check rather than a necessity.

%% file: sections/06_next_steps.tex
\section{Next steps and release}\label{sec:next}

\paragraph{More sensors and units.}
A fourth family, the XELA uSkin \citep{tomo2018new}, records three-axis magnetic taxels instead of images, so the same
contacts will link optical and non-optical sensing. Every family will be recorded on two units, to measure and
learn the variation between units. DIGIT needs this most, since its units are known to generalise poorly to one another
\citep{10851808}; for GelSight Mini and Daimon, whose units are much more consistent, the second unit is probably not
necessary but is recorded as a check.

\paragraph{More geometry.}
The indenter set grows to about 50 shapes, recorded on one unit of each family, with a fixed subset on the other
units. Our first results favour this breadth: holding out whole contact points barely changed the read-outs, while
unseen indenters and unseen units did, so fewer contact points per indenter will buy more indenters.

\paragraph{Manipulation-like force histories.}
The current trials load, hold and unload. Because the controller tracks any force history, the same rig can reproduce
the force histories a fingertip sees in manipulation, with every trajectory still commanded identically on all
sensors and kept inside the common envelope: grip force followed by load force, as when lifting an object
\citep{johansson1984roles}; tangential force changing direction at constant grip, as when carrying it; grip force
cycling, as in regrasping or a handover \citep{marra2026temporal}; and faster loading and release.

\paragraph{Release.}
The project repository (\url{https://github.com/hsp-iit/rosetta-touch}) describes the dataset and how to obtain it. The data recorded so far
are available on request; the complete dataset will be released there under a CC~BY~4.0 licence once collection ends,
with the calibration, the pairing manifests, loaders and the encoder trained on the data.